\pdfoutput=1
\documentclass[11pt]{article}

\usepackage[preprint]{acl}

\usepackage{times}
\usepackage{latexsym}

\usepackage[T1]{fontenc}
\usepackage[utf8]{inputenc}

\usepackage{microtype}

\usepackage{inconsolata}

\usepackage{graphicx}

\usepackage{times}  % DO NOT CHANGE THIS
\usepackage{helvet}  % DO NOT CHANGE THIS
\usepackage{graphicx} % DO NOT CHANGE THIS
\usepackage{multirow} 
\usepackage{url}            % simple URL typesetting
\usepackage{booktabs}       % professional-quality tables
\usepackage{amsfonts}       % blackboard math symbols
\usepackage{nicefrac}       % compact symbols for 1/2, etc.
\usepackage{microtype}      % microtypography
\usepackage{xcolor}
\usepackage{amsmath}
\usepackage{amssymb}
\usepackage{mathtools}
\usepackage{amsthm}
\usepackage{xspace}
\usepackage{subcaption}
\usepackage{pifont} % 用于 \ding{51} 和 \ding{56} (打钩和叉)
\usepackage{threeparttable} % 用于表格下方的注释（tablenotes）
\usepackage{array} % 允许在表格中使用自定义列格式

\newcommand{\I}{\mathcal{I}}

\newcommand{\T}{\mathcal{T}}
\newcommand{\D}{\mathcal{D}}
\newcommand{\M}{\mathcal{M}}

\newcommand{\alg}{\textsc{SafeClip}\xspace}
\newcommand{\algo}{\textsc{OTCClip}\xspace}

\usepackage{algorithm}
\usepackage{algpseudocode} % for pseudocode functionality

\usepackage{newfloat}
\usepackage{listings}

\title{Pre-training CLIP against Data Poisoning with Optimal Transport-based Matching and Alignment}

\author{
	Tong Zhang$^{1,}$\thanks{ Equal contribution},  Kuofeng Gao$^{2,}$\footnotemark[1], Jiawang Bai$^{2,}$\footnotemark[1], Leo Yu Zhang$^{3}$, Xin Yin$^{1}$ \\
    \bf Zonghui Wang$^{1,}$\thanks{ Corresponding authors}, \bf Shouling Ji$^{1,}$\footnotemark[2], \bf Wenzhi Chen$^{1}$ \\
$^1$Zhejiang University \quad 
$^2$Tsinghua University \quad
$^3$Griffith University \\
\texttt{tz21@zju.edu.cn, gkf21@mails.tsinghua.edu.cn, bjw19@tsinghua.org.cn,} \\
    \texttt{leo.zhang@griffith.edu.au, \{xyin, wangzonghui, sji, chenwz\}@zju.edu.cn} \\
}	

\begin{document}
\maketitle
\begin{abstract}
Recent studies have shown that Contrastive Language-Image Pre-training (CLIP) models are threatened by targeted data poisoning and backdoor attacks due to massive training image-caption pairs crawled from the Internet. Previous defense methods correct poisoned image-caption pairs by matching a new caption for each image. However, the matching process relies solely on the global representations of images and captions, overlooking fine-grained features of visual and textual features. It may introduce incorrect image-caption pairs and harm the CLIP pre-training. To address their limitations, we propose an Optimal Transport-based framework to reconstruct image-caption pairs, named \algo. We propose a new optimal transport-based distance measure between fine-grained visual and textual feature sets and re-assign new captions based on the proposed optimal transport distance. Additionally, to further reduce the negative impact of mismatched pairs, we encourage the inter- and intra-modality fine-grained alignment by employing optimal transport-based objective functions.
Our experiments demonstrate that \algo can successfully decrease the attack success rates of poisoning attacks. Also, compared to previous methods, \algo significantly improves CLIP's zero-shot and linear probing performance trained on poisoned datasets.  

% \thanks{Code available at \href{https://github.com/ShiyuXiang77/ALRPHFS}{https://github.com/ShiyuXiang77/ALRPHFS}}%
\end{abstract}

\section{Introduction}
\looseness=-1
\begin{figure}[t!]
    \begin{center}
        \includegraphics[width=0.95\linewidth]{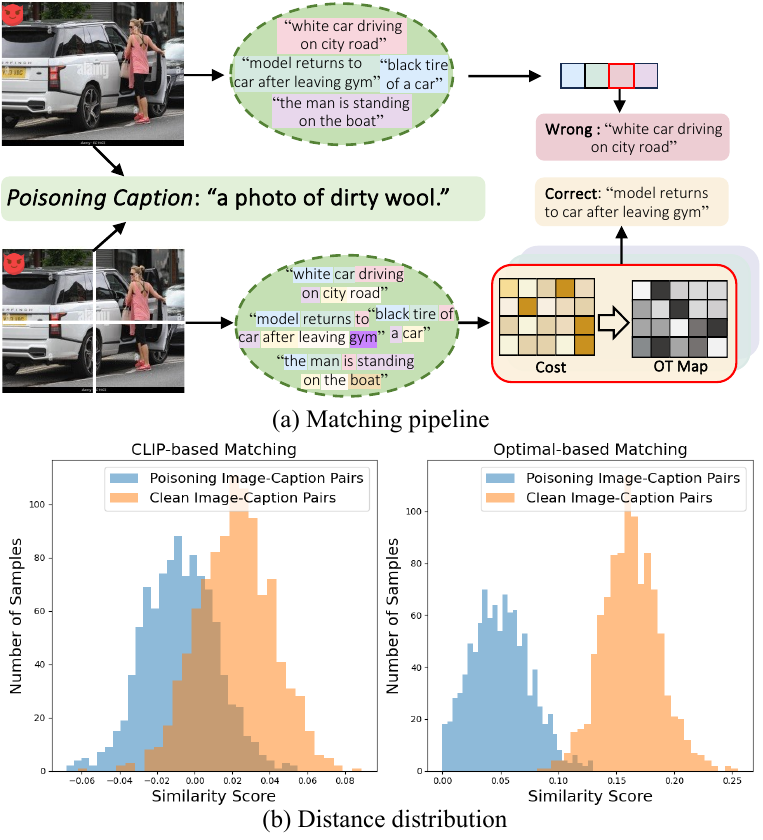}
    \end{center}
    \vspace{-2mm}
    \caption{(a) Previous methods use global features of the CLIP model, while we employ a fine-grained optimal transport method. (b) Compared to CLIP-based matching, optimal transport-based fine-grained matching is robust for distinguishing poisoned data.}
    \label{fig:fig_1}
    \vspace{-2mm}
\end{figure}
Contrastive Language-Image Pre-training (CLIP) models have demonstrated remarkable zero-shot performance across diverse domains, leveraging millions or billions of training samples from the Internet~\cite{radford2021learning,jia2021scaling}. As CLIP's large-scale pre-training data is often crawled online, attackers can inject malicious examples into the training set to alter predictions at test time. Recent research has shown that multimodal models are vulnerable to data poisoning and backdoor attacks~\cite{liang2025revisiting,xiang-etal-2025-beyond,liang2022large,gao2023backdoorhash,xiang2025alrphfs}. Specifically, CLIP are more susceptible to both targeted data poisoning attacks (TDPAs) and backdoor attacks (BAs), where the insertion of adversarial triggers into as little as 0.01\% of the pre-training data can reliably induce targeted misclassification, while TDPAs are even more effective, requiring only 0.0001\% poisoned data~\cite{carlini2024poisoning,carlini2021poisoning, yang2023robust}.

An effective defense method is crucial to mitigate the impact of poisoned image-caption pairs during pre-training. RoCLIP~\cite{yang2023robust} disrupts the malicious link between poisoned images and captions by matching each image representation with its most similar caption from a random pool.
\alg~\cite{yang2024better} avoids misleading information by employing cross-modal alignment only on identified clean datasets. The ability to distinguish poisoned data is highly dependent on the matching method used to identify or correct the data while the model is not yet fully trained. As shown in Figure~\ref{fig:fig_1}, previous methods uses CLIP-based semantic matching to differentiate poisoned image-caption pairs face challenges. This is because global features focus on overall semantics, which means that subtle yet indicate poisoning or inconsistencies within image-caption pairs are likely to be missed. Hence, identification and correction result in a suboptimal solution, which ultimately causes the model to overfit to poisoned data.

In this work, we leverage fine-grained features to address the limitations mentioned above and enhance the model's generalization capability. To achieve this, we introduce the optimal transport framework \algo, designed to disrupt the association between poisoning image-caption pairs during pre-training.  We consider the fine-grained feature similarity measure as an optimal transportation problem to reconstruct and align the image-caption pairs, which aims to transport a collection of contextual patches in an image to the ones in another contextualized token sequence in a caption. \algo first employs optimal transport-based matching, using the transport matrix as weights to effectively capture relationships between different regions of image patches and caption tokens. This approach improves the ability of the model to distinguish poisoned data, as shown in Figure~\ref{fig:fig_1}. 

However, it is challenging to correct all poisoned data solely through optimal transport-based matching. Therefore, we propose a fine-grained alignment module to further enhance resilience for poisoned data. \algo treats the alignment of images and captions obtained from optimal transport matching as a distribution transport optimization task to better associate image patches and caption tokens. Optimal transport assigns greater weights in highly similar regions of image-caption pairs and smaller weights to less similar regions, which reduces the risk of introducing errors from unmatched pairs during pre-training. In addition, intrinsic relationships within each modality are crucial and are not affected by cross-modal poisoning. Hence, we separately employ the intra-modality fine-grained alignment for image patches and caption tokens to further against data poisoning.

We conduct extensive experiments on multiple image-caption datasets, showing that \algo effectively reduces attack success rates to 0\% in most cases. Additionally, we observe improvements in CLIP's zero-shot and linear probing performance. \looseness=-1

\section{Related Work}
\label{sec:related_work}
\subsection{Protecting CLIP Against TDPA and BA}
CLIP is vulnerable to targeted data poisoning attacks (TDPAs) and backdoor attacks (BAs) \citep{carlini2021poisoning, yang2023data}. TDPAs manipulate a small portion of training data to mislead model into misclassifying specific examples, while BAs embed visible or invisible triggers (e.g., noise or deformations) induce misclassification of test images containing the same trigger \citep{chen2017targeted, gu2017badnets, nguyen2021wanet}.

Effective defense methods have been proposed recently, which can be divided into four, including against backdoor/poisoning pre-training~\cite{yang2023robust,yang2024better}, fine-tuning the backdoored CLIP~\cite{bansal2023cleanclip,kuang2024adversarial,xuncleanerclip}, using trigger inversion~\cite{sur2023tijo,feng2023detecting}, and backdoor detection~\cite{niu2024bdetclip,huang2025detecting}. Remarkably, research has shown that adding a trigger to just 0.01\% of pre-training data can cause misclassification~\cite{bansal2023cleanclip}, while TDPAs are even more effective, requiring only 0.0001\% poisoned data~\cite{yang2023robust,yang2024better}. Current defenses for CLIP remain limited against these attacks.

RoCLIP \citep{yang2023robust} against data poisoning and backdoor attacks by augmenting image-caption pairs and matching them with nearest-neighbor captions from a pool in the pre-training. However, it overlooks local features and relies solely on global semantics, which can introduce matching errors and degrade performance. \alg \citep{yang2024better} avoid involving the misleading information by employing cross-modal alignment on clean datasets. \alg  first distinguish safe from risky data pairs by overall semantic features between image and caption datasets. \alg only apply cross-modal alignment on clean samples, harming the model's performance. For example, with a poisoning rate of 0.5\%, more than 70\% of clean data is classified into the harmful dataset, solely by applying self-modal feature alignment, which harms the model's performance.

\subsection{Vision-Language Feature Alignment}
% Fine-grained feature alignment is essential for providing accurate supervisory signals and improving model performance. FILIP \cite{yao2021filip} enhances alignment by using token-wise maximum similarity between visual and textual tokens. Other methods, such as OSCAR \cite{li2020oscar}, VinVL \cite{zhang2021vinvl}, MVPTR \cite{li2022mvp}, and X-VLM \cite{zeng2021multi}, focus on multi-level semantic alignment. OSCAR introduces multi-level semantics by capturing object region features and tags, while VinVL refines visual features with an improved object-attribute detector. MVPTR and X-VLM extend multi-level semantics across both visual and textual modalities, with MVPTR modeling object-tag alignment and phrase structure, and X-VLM aligning visual concepts with textual descriptions. PyramidCLIP \cite{gao2022pyramidclip} combines three visual and three linguistic representations to compute multiple contrastive loss terms, supporting multi-level alignment. Collectively, these approaches show that fine-grained features enhance image-caption alignment and boost resilience to perturbations. 
Fine-grained feature alignment is key to providing accurate supervision and improving model performance. FILIP \cite{yao2021filip} achieves this via token-wise maximum similarity between visual and textual tokens. Other methods, such as OSCAR \cite{li2020oscar}, VinVL \cite{zhang2021vinvl}, MVPTR \cite{li2022mvp}, and X-VLM \cite{zeng2021multi}, focus on multi-level semantic alignment. OSCAR introduces multi-level semantics by capturing object region features and tags, while VinVL refines visual features with an improved object-attribute detector. MVPTR and X-VLM extend multi-level semantics across both visual and textual modalities, with MVPTR modeling object-tag alignment and phrase structure, and X-VLM aligning visual concepts with textual descriptions. PyramidCLIP \cite{gao2022pyramidclip} combines three visual and three linguistic representations to compute multiple contrastive loss terms, supporting multi-level alignment. Collectively, these approaches show that fine-grained features enhance image-caption alignment and boost resilience to perturbations.

\section{Preliminary}%\vspace{-2mm}
\subsection{Contrastive Language-Image Pre-training (CLIP)}\label{sec:pre_clip}

Typically, CLIP employs two main components: an image encoder $E_I$ and a text encoder $E_T$. Given a dataset $\D$ consisting of image-caption pairs $(\mathbf{X}_i, \mathbf{Y}_i)$, where $\mathbf{X}_i$ represents the image, and $\mathbf{Y}_i$ represents the corresponding caption.   
When the image $\mathbf{X}_i$ is input into the image encoder $E_I$, it is first transformed into spatial feature representations $f_i^s \in \mathbb{R}^{h \times w\times d} $, then condensed into a global feature vector $f_i^g \in \mathbb{R}^{d} $. These spatial features can be represented as \( f_i^s = \{ z^s_{i,1}, z^s_{i,2}, \cdots, z^s_{i,{h \times w}} \} \), where each \( z^s_{i,j} \in \mathbb{R}^d \) (for \( j = 1, 2, \dots, h \times w \)) is a feature vector corresponding to a spatial location in the image. The spatial features are then condensed into a global feature vector \( f_i^g \in \mathbb{R}^d \). Here, $h$ and $w$ denote the height and width of the feature map, while $d$ represents the dimensionality of each feature at a given spatial location. Similarly, the text $\mathbf{Y}_i$ is encoded into the text encoder $E_T$ to produce token sequence features $y_i^s \in \mathbb{R}^{l\times d} $, which are further aggregated into a global feature $y_i^g \in \mathbb{R}^{d} $. These token sequence features are represented as \( y_i^s = \{ \hat{z}^s_{i,1}, \hat{z}^s_{i,2}, \cdots, \hat{z}^s_{i,l} \} \), where each \( \hat{z}^s_{i,j} \in \mathbb{R}^d \) (for \( j = 1, 2, \dots, l \)) is a token vector corresponding to a position in the caption. Here, $l$ denotes the length of the token sequence feature, while $d$ represents the dimensionality.
To enable multi-modal interaction, CLIP employs the InfoNCE loss during training. This loss function encourages the alignment of representations from each image-caption pair while separating those of non-paired images and captions within the same mini-batch. The quality of the learned representations is assessed using zero-shot and linear probe classification; details of these evaluation protocols are provided in Section~\ref{sec:add_expr}.
\looseness=-1

\subsection{Threat Model}
\label{sec:pre_attack}
\noindent\textbf{Adversary capabilities.}
Recent research~\cite{yang2023robust,yang2024better,bai2024badclip,liang2024badclip} has revealed the serious backdoor vulnerability of CLIP. We adopt the poisoning-based threat model from previous works~\cite{yang2023robust,yang2024better}, where the adversary injects a set of poisoning image-caption pairs into the pre-training data. In this scenario, attackers can only manipulate poisoned data, unlike other works~\cite{bai2024badclip,liang2024badclip}, which assume attackers modify the training process. Let $D_{poi} = {(\mathbf{X}_i, \mathbf{Y}_{poi(i)})| \mathbf{X}_i\in\I_{i}, \mathbf{Y}_{poi(i)} \in \T{adv}}$ denote the injected poisoning pairs, where $\D_{poi} \subset \D$. Here, $\T_{adv}$ is the set of adversarial captions related to the adversarial label $\mathbf{Y}_{adv}$. There are two ways to generate adversarial captions. On one hand, the adversary can construct an adversarial caption by searching for some captions containing the adversarial label. Alternatively, the adversarial can utilize CLIP's 80 prompt-engineered text descriptions \cite{radford2021learning,yang2023robust,zhou2022conditional} to generate captions for the adversarial label. Besides, the adversaries have knowledge of the model's architecture, the training algorithm, and the hyperparameters but cannot directly the alter training process.

% We assume that the adversary has limited control over the pre-training data, and can inject a small number of poisoned examples ($\le 0.05\%$ of the dataset size for TDPA and $\le 0.15\%$ of the dataset size for BA) into the training dataset. Adversary also has the knowledge of the model structure, the training algorithm, and the hyperparameter used by their victim, but they cannot modify the training process directly.\looseness=-1

\noindent\textbf{Adversary objective.} Targeted data poisoning attacks aim to misclassify a particular test example, $\mathbf{X}_i$, as $\mathbf{Y}_{adv}$. Hence, 
$D_{poi}=\{(\mathbf{X}_i, \mathbf{Y}_{poi(i)})|\mathbf{Y}_{poi(i)}\in\T_{adv}\}$. Backdoor attacks introduce a trigger patch to a set of poisoned images. The goal is to misclassify any test examples with the trigger patch, $\mathbf{X}_i \oplus \text{patch}$, as $\mathbf{Y}_{adv}$. Hence, $D_{poi}=\{(\mathbf{X}_i\oplus \text{patch},\mathbf{Y}_{poi(i)})| \mathbf{X}_i\in\I, \mathbf{Y}_{poi(i)}\in\T_{adv}\}$. In contrast to targeted data poisoning attacks, which target a particular test example, backdoor attacks inject \textit{random} images with the backdoor trigger, paired with the adversarial captions.

\section{Method}

In this section, we first introduce the foundational concepts of optimal transport and describe how the fine-grained matching problem can be modeled in an optimal transport framework. Next, we explain the fine-grained alignment module and provide the implementation details for training and inference.

\begin{figure*}[t!]
    \begin{center}
        \includegraphics[width=1\linewidth]{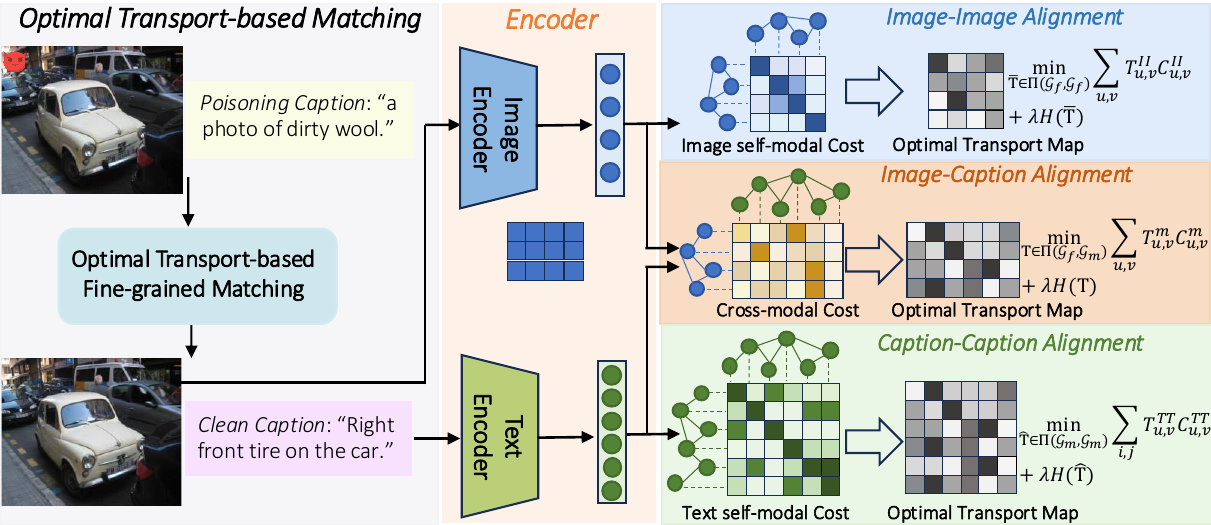}
    \end{center}
    \vspace{-2mm}
    \caption{Illustration of \algo for defending CLIP during pre-training. Given image-caption pairs, \algo first applies optimal transport matching to break the association between poisoning images and captions, reconstructing new image-caption pairs. These reconstructed pairs are then fed into the optimal transport-based inter-modality module to better align fine-grained features and reduce the negative impact of mismatches.  Reconstructed pairs also are fed into the optimal transport-based intra-modality alignment module to capture the  intrinsic relationships of each modality. Additionally, reconstructed data use CLIP’s  InfoNCE loss to achieve alignment of semantics.}
    \label{fig:framework}
\end{figure*}

\subsection{The Definition Of Optimal Transport}

\noindent\textbf{Defining Source And Target Distributions.} First, we define two pivotal distributions within the optimal transport framework~\cite{pramanick2023volta,chang2022unified}: the source distribution \( \mathbf{K} = (k_1, k_2, \cdots, k_n) \) and the target distribution \( \mathbf{Q} = (q_1, q_2, \cdots, q_m) \). These distributions correspond to the starting and ending points of the transportation process.

\noindent\textbf{Transportation matrix T.} 
The transportation plan is described by a matrix \( \mathbf{T} = [T_{uv}] \) of size \( n \times m \). Each element \( T_{uv} \) represents the amount of resource transported from the \( u \)-th source in \( \mathbf{K} \) to \( v \)-th target in \( \mathbf{Q} \). This matrix outlines the optimal transportation strategy, aligning the two distributions while minimizing  total cost~\cite{chen2020graph}.

In the optimal transport framework, the matrix \( \mathbf{T} \) must meet specific constraints to ensure an effective transportation plan~\cite{chen2020graph,li2024unsupervised}. The Marginal Constraints are given by \( \sum_{v=1}^{m} T_{uv} = k_u \) for \( u = \{1, \ldots, n\} \) and \( \sum_{u=1}^{n} T_{uv} = q_v \) for \( v = \{1, \cdots, m \} \). These constraints require that the total transported amount from each source \( u \) and to each target \( v \) matches the respective supply \( k_u \) and demand \( q_v \). The Non-Negativity Constraint is \( T_{uv} \geq 0 \) for all \( u \) and \( v\), ensuring all transport amounts \( T_{uv} \) are non-negative, which reflects the practical impossibility of negative transportation.

% In the optimal transport framework, the matrix \( \mathbf{T} \) must meet specific constraints to ensure an effective transportation plan~\cite{chen2020graph}. The Marginal Constraints are:
% \begin{align}
% \vspace{-4mm}
% \sum_{j=1}^{m} T_{ij} &= k_i \quad \forall i \in \{1, \ldots, n\}, \\
% \vspace{-2mm}
% \sum_{i=1}^{n} T_{ij} &= q_j \quad \forall j \in \{1, \ldots, m\}.
% \vspace{-4mm}
% \end{align}
% These constraints require that the total transported amount from each source \( i \) and to each target \( j \) matches the respective supply \( k_i \) and demand \( q_j \).
% %
% Additionally, the Non-Negativity Constraint is:
% \begin{equation}
% T_{ij} \geq 0 \quad \forall i \in \{1, \ldots, n\}, \forall j \in \{1, \ldots, m\}.
% \end{equation}
% This condition ensures all transport amounts \( T_{ij} \) are non-negative, reflecting the practical impossibility of negative transportation.

\noindent\textbf{Modeling the optimal transport problem.} With the aforementioned definitions and constraints established, the Optimal Transport problem can be formulated as follows:
\begin{equation}
\small
OT(\mathbf{K}, \mathbf{Q}, \mathbf{C}) 
= \min_{\mathbf{T} \in \Pi(\mathbf{K}, \mathbf{Q})} \sum_{u=1}^n \sum_{v=1}^m T_{uv} \cdot C_{uv},
\label{eq:wd}
\end{equation}
where $C$ denotes the cost matrix, with each element \( C_{uv} \) representing the cost of transporting a unit from source \( k_u \) to target \( q_v \). The matrix \( \mathbf{T} \) signifies the transportation scheme, while \( \Pi(\mathbf{K}, \mathbf{Q}) \) encompasses all feasible transportation schemes that satisfy the marginal constraints.

% Next, Sinkhorn distance is used in Optimal Transport (OT) for its effectiveness in high-dimensional spaces. Traditional OT approaches, based on linear programming, face challenges with computational intensity and scaling with data dimensionality. In contrast, the Sinkhorn distance applies entropy regularization to the OT calculation, enhancing tractability and differentiability. This approach uses a regularization parameter \( \lambda \), which balances accuracy and computational efficiency. Higher \( \lambda \) values lead to results closer to traditional OT but at increased computational costs, while lower values of \( \lambda \) expedite calculations at the expense of some bias. Therefore, the Sinkhorn distance, often computed using the Sinkhorn-Knopp algorithm, presents a more feasible solution for OT in machine-learning scenarios that demand scalability and stability in computations.

To handle high-dimensional spaces effectively, the Sinkhorn distance is used in Optimal Transport (OT) \cite{distances2013lightspeed}. Traditional OT methods, which rely on linear programming, struggle with computational demands and scalability issues. In contrast, the Sinkhorn distance incorporates entropy regularization into the OT calculation, improving both tractability and differentiability. Consequently, the Sinkhorn Optimization Process can be defined as:
\begin{equation}
\label{eq:ot_}
\small
% D^{P}(\boldsymbol{\mathcal{G}_f},\boldsymbol{\mathcal{G}_p})= \min_{\mathbf{T} \in \Pi(\boldsymbol{\mathcal{G}_f},\boldsymbol{\mathcal{G}_p})} \sum_{i,j} T^P_{ij}C^P_{ij} + 
M(\mathbf{K}, \mathbf{Q}, \mathbf{C}) 
=\min_{\mathbf{T} \in \Pi(\mathbf{K}, \mathbf{Q})}\sum_{u=1}^n \sum_{v=1}^m T_{uv} \cdot C_{uv}+\lambda H(\mathbf{T}),
% \vspace{-2mm}
\end{equation}
where \( H(\mathbf{T}) \) is the entropy of the transport matrix, which introduces regularization to ensure numerical stability and efficient computation, and \( \lambda \) is a hyper-parameter that balances accuracy and computational efficiency. Higher \( \lambda \) values yield results closer to traditional OT but increase computational costs, while lower values of \( \lambda \)  speed up calculations at the cost of some bias. The Sinkhorn algorithm iteratively normalizes the rows and columns of the transport matrix to satisfy the marginal constraints while minimizing the regularized objective function~\cite{distances2013lightspeed}.

%%%%%%%%%% Optimal Transport Matching %%%%%%%%%%
%%%%%%%%%% Optimal Transport Matching %%%%%%%%%%
%%%%%%%%%% Optimal Transport Matching %%%%%%%%%%
% For image-caption pairs, we achieve an inter-modality fine-grained matching for them by calculating the matching degree between images and caption.

% Instead of matching every image with its corresponding caption

\subsection{Optimal Transport-based Matching}
\label{sec:inter_match}
Previous methods~\cite{yang2023robust,yang2024better} use the global feature to identify the poisoning data. However, global features tend to emphasize only the most prominent or frequent characteristics in the data, primarily capturing dominant semantic information while overlooking finer details. The global feature focus on overall semantics means that subtle yet important cues, especially those that may indicate poisoning or inconsistencies within image-caption pairs, are likely to be missed. To address this issue, we employ optimal transport into fine-grained matching between images and captions. Given an image with spatial features $f^s_i$, our aim is to find the most matching caption from a randomly sampled pool of captions with fine-grained features $\mathcal{P}^s = \{y^s_{p(i)}\}_{i=1}^P$. Given the definition of optimal transport, we define the fine-grained feature set \( f_i^s = \{ z^s_{i,1}, z^s_{i,2}, \cdots, z^s_{i,{h \times w}} \} \) as a distribution of patch-level features $\boldsymbol{\mathcal{G}_f}$. 
Similarly, we define the set of token sequence features \( y_{p(j)}^s = \{ \hat{z}^s_{j,1}, \hat{z}^s_{j,2}, \cdots, \hat{z}^s_{j,l} \} \) in the caption pool as the distribution of token-level features $\boldsymbol{\mathcal{G}_p}$.

To perform the fine-grained matching, we first compute a similarity matrix $ S^{P}=f_i^s \odot y_{p(j)}^s$ between image patches and caption tokens. Here, $\odot$ represents the Hadamard product and $S^P\in \mathbb{R}^{ hw\times l}$. Each position in the similarity matrix focuses only on local features between image patches and caption tokens. Therefore, the similarity matrix cannot effectively represent the global matching degree between the image and caption. In the optimal transport, the overall matching cost $\sum_{u,v} T_{uv}C^{P}_{uv}$ is calculated by the product sum of the transportation matrix $T$ and the cost matrix $C$, the cost matrix is defined as $C^{P} = 1 - S^{P}$. By optimizing the transport plan, the transport matrix determines how to match image patches and caption tokens at the minimum cost. Therefore, optimal transport can measure the degree of overall matching between the image patches and the caption tokens from a global perspective. Then, the overall matching score $M$ between a given image and any caption sequence in the pool can be calculated as follows:
% {\color{blue}Accordingly, the cost matrix is defined as $C^{P} = 1 - S^{P}$, where $S^{P}=\boldsymbol{\mathcal{G}_f} \odot \boldsymbol{\mathcal{G}_p}$ is the similarity matrix. Then, the overall matching score $M$ between a given image and any caption sequence in the pool can be calculated as follows:}
% To perform the fine-grained matching, we first compute a similarity matrix $S^{P}=\boldsymbol{\mathcal{G}_f} \odot \boldsymbol{\mathcal{G}_p}$ between image patches and caption tokens. Here, $\odot$ represents the Hadamard product and $S^P(f_i^s,y_{p(j)}^s)\in \mathbb{R}^{ hw\times l}$. Each position in the similarity matrix focuses only on local features between image patches and caption tokens. Therefore,  the similarity matrix cannot effectively represent the global matching degree between the image and caption. In the optimal transport, the overall matching cost $\sum_{i,j} T_{ij}C^{P}_{ij}$ is calculated by the product sum of the transportation matrix $T$ and the cost matrix $C$. By optimizing the transport plan, the transport matrix determines how to match image patches and caption tokens at the minimum cost. Therefore, optimal transport can measure the degree of overall matching between the image patches and the caption tokens from a global perspective. Specifically, we define a cost matrix as $C^{P} = 1 - S^{P}$. Then, the overall matching score $M$ between a given image and any caption sequence in the pool can be calculated as follows:
\begin{equation}
\label{eq:matching}
M = \min_{\mathbf{T} \in \Pi(\boldsymbol{\mathcal{G}_f},\boldsymbol{\mathcal{G}_{p}})} \sum_{u,v} T_{uv}C^{P}_{uv} + \lambda H(\mathbf{T}),
\end{equation}
where the optimization algorithm for the transport matrix is outlined in Algorithm~\ref{alg:sinkhorn}. This matrix optimizes by identifying the best associations between image patches and token sequences, reducing the risk of mismatches. Since a lower optimal transport matching score indicates greater similarity between image-caption pairs, we redefine $\hat{M} = 1 - M$ to align with CLIP’s concept of similarity in matching. 
For different pixel-level feature sets and token-sequence feature sets, we have different representations for the distribution of patch-level features $\boldsymbol{\mathcal{G}_f}$, token-level features $\boldsymbol{\mathcal{G}_p}$, the similarity matrix $S^{P}$, and the transportation matrix $T$. To simplify the notation, we omit the corresponding subscripts.

Given $N$ image within a mini-batch, we compute the similarity score between each image and every caption in the caption pool, resulting in $N$ similarity matrix $\M =\{\hat{M}^P_i\}^N_{i=1} \in \mathbb{R}^{N\times P}$. Next, for each image features, we select the most matching caption feature from the pool $\mathcal{P}^s$ based on the similarity matrix. For the $j$-th image feature, the selected caption feature is as follows:
\begin{equation}
  \label{eq:index}
{y}^{s}_{m(j)}=y_{p}\left[\arg \max _{1 \leq p \leq P} \hat{M}_j^P[p]\right].  
\end{equation}
Through above operations, we can obtain the matched caption $y_{m(j)}^s$ in the pool that is most similar to $f^s_j$, resulting in the matching fine-grained feature $\{f_j^s,y_{m(j)}^s\}_{j=1}^N$ within a mini-batch. Similarly, we can obtain the global feature $\{f_j^g,y_{m(j)}^g\}_{j=1}^N$ for the matched image-caption pairs. Therefore, we can break the  poisoning data to prevent it from being used during pre-training.

% \begin{algorithm}[t]
% \caption{Sinkhorn Iteration for Optimal Transport}
% \label{alg:sinkhorn}
% \begin{algorithmic}[1]
% \Require $C$: cost matrix, $P$: number of caption pool, $h \times w$: number of spatial image features, $l$: length of caption, $\beta$: scaling parameter
% \Ensure $T$: transport matrix

% \State $\sigma \gets \text{ones\_like}(P, h \times w, 1) / m$
% \State $T \gets \text{ones\_like}(P, l, h \times w)$
% \State $A \gets \exp(-(\text{clamp}(C, \max(10 \cdot \beta))) / {\beta})$
% \For{$i = 1$ \textbf{to} $100$}
%     % \State $r_0 \gets r$
%     \State $\delta \gets 1 / 1 / {n \cdot \sum(Q \cdot \sigma, \text{ axis}=2)}$
%     \State $a = \sum(Q \cdot \delta, \text{ axis}=2)$
%     \State $\sigma = 1 / {m \times a}$
%     \State $T \gets \delta \times Q \times K$
% \EndFor
% \State \Return $T$
% \end{algorithmic}
% \end{algorithm}

\subsection{Fine-grained Alignment}
Through optimal transport-based matching, we obtain the global feature $\{f_j^g,y_{m(j)}^g\}_{j=1}^N$ of the matched image-caption pairs. To facilitate multi-modal interaction, we first use the CLIP loss for optimization as follows: 
\begin{equation}
\label{eq:clip}
\small
\begin{split}
    \mathcal{L}_c = &-\frac{1}{2N} \sum_{i=1}^N \log \left [ \frac{\exp\left(\left<f_i^g,y_{m(i)}^g\right>/\tau\right)}{\sum_{j=1}^N \exp\left(\left<f_i^g,y_{m(j)}^g\right>/\tau\right) } \right] \\
    &-\frac{1}{2N} \sum_{j=1}^N \log \left [ \frac{\exp\left(\left<f_j^g,y_{m(j)}^g\right>/\tau\right)}{\sum_{i=1}^N \exp\left(\left<f_i^g,y_{m(j)}^g\right>/\tau\right) } \right],
\end{split}
\end{equation}
where $\tau$ is the temperature coefficient in CLIP.

% Based on the optimal transport-based matching image-caption pairs, we propose three loss functions for alignment between them, which can effectively defend targeted data poisoning and backdoor attacks.
% % Building on the caption pool for each image, we extract the global features $\mathcal{P}^g = \{y^g_{p(i)}\}_{i=1}^P$.
% First, given fine-grained matched image-caption pairs obtained in Section \ref{sec:inter_match}, we extract the global features $\{f_j^g, y_{m(j)}^g\}_{j=1}^N$ for them.
% To enable multi-modal interaction, we use the CLIP loss for optimization as follows: 
% \begin{equation}
% \vspace{-2mm}
% \label{eq:clip}
% \small
% \begin{split}
%     \mathcal{L}_c = &-\frac{1}{2N} \sum_{j=1}^N \log \left [ \frac{\exp\left(\left<f_j^g,y_{m(j)}^g\right>/\tau\right)}{\sum_{k=1}^N \exp\left(\left<f_j^g,y_{m(k)}^g\right>/\tau\right) } \right] \\
%     &-\frac{1}{2N} \sum_{k=1}^N \log \left [ \frac{\exp\left(\left<f_k^g,y_{m(k)}^g\right>/\tau\right)}{\sum_{j=1}^N \exp\left(\left<f_j^g,y_{m(k)}^g\right>/\tau\right) } \right].
% \end{split}
% \vspace{-2mm}
% \end{equation}

% In addition, we also introduce inter-modality structure alignment to better align the structural information of image-caption pairs. To further alignment between target image sub-graph and text sub-graph, we use the CLIP loss for optimization.  Next, the clean global text-caption pairs features will be trained in a mini-batch as follows:

\noindent\textbf{Inter-modality Fine-grained Alignment.} 
\label{sec:inter}
% However, relying solely on the CLIP semantic loss can lead the model to focus too much on global features while overlooking fine-grained details, which can introduce erroneous information during matching. To address this challenge, we propose fine-grained alignment across different modalities.
In addition to the CLIP semantic loss, which focuses on global feature alignment, we further propose a fine-grained feature alignment loss across different modalities.
Similarly Eq~\ref{eq:matching}, for any single matched pair $\{f_j^s, y_{m(j)}^s\}$ within the set $\{f_i^s, y_{m(i)}^s\}_{i=1}^N$, we define the distribution of patch-level features of images and token-level features of matched captions $\boldsymbol{\mathcal{G}_f}$ and $\boldsymbol{\mathcal{G}_{m}}$, respectively.
Then, we define the cost matrix $C^{m} = 1 - S^{m}$, where $S^{m}$ denotes the similarity matrix between image patches and caption tokens within an image-caption pair. The loss for inter-modality fine-grained alignment can be defined as the optimal transport problem as follows:
\begin{equation}
\small
\label{eq:inter_modal_loss}
\mathcal{L}^a = \min_{\mathbf{T} \in \Pi(\boldsymbol{\mathcal{G}_f},\boldsymbol{\mathcal{G}_{m}})} \sum_{u,v} T^{m}_{uv}C^{m}_{uv} + \lambda H(\mathbf{T}).
\end{equation}
For $N$ image-caption pairs in a mini-batch, we compute the loss for each pair, resulting in $N$ losses $\{\mathcal{L}_i^a\}_{i=1}^{N}$. The total inter-modality fine-grained alignment loss is the sum of all individual losses as $\mathcal{L}_{IM} = \sum_{i=1}^N \mathcal{L}^a_i$. It can enhance the alignment between matched image patches and caption tokens while simultaneously maximizing the separation between non-matching ones. During optimization, the transport matrix assigns larger weights to image patches and caption tokens with higher similarity and smaller weights to those with lower similarity. Therefore, the model effectively alleviates the risk of being negatively affected by irrelevant information during training by prioritizing the high-similarity image patches and caption tokens. 
This is achieved through the optimization of the transport matrix, as outlined in Algorithm~\ref{alg:sinkhorn}.

\noindent\textbf{Intra-modality Fine-grained Alignment.} 
\label{sec:intra}
While inter-modal fine-grained alignment can improve the feature correspondence between image patches and text tokens, it is not sufficient to fully resolve the model's confusion during training. For example, in an image containing multiple instances of the same object (\textit{e.g.}, multiple ``tires"), inter-modal fine-grained alignment will treat all these instances as identical, failing to capture the different intra-modal relationships like \textit{``Right front on the car"}.

To address this limitation, we propose an intra-modal fine-grained alignment approach. Specifically, given two distributions $\boldsymbol{\mathcal{G}_f}$ and $\boldsymbol{\mathcal{G}_{m}}$ introduced in Eq~\ref{eq:inter_modal_loss}, we first compute the similarity matrix for text-to-text pairs, denoted as \(S^{TT} \in \mathbb{R}^{hw \times hw}\), and for image-to-image pairs, denoted as \(S^{II} \in \mathbb{R}^{l \times l}\), similar to Section~\ref{sec:inter_match}.
We then derive the cost matrices \(T^{II}\) and \(T^{TT}\) for each distribution. The loss function for intra-modality fine-grained alignment is defined as follows:
\begin{equation}
\label{eq:str}
\small
\begin{aligned}
\mathcal{L}^s & =  \min_{\bar{\mathbf{T}} \in \Pi(\boldsymbol{\mathcal{G}_f},\boldsymbol{\mathcal{G}_f})} \sum_{u,v} T^{II}_{uv}C^{II}_{uv}  + \lambda H(\bar{\mathbf{T}}) \\
& \quad + \min_{\hat{\mathbf{T}} \in \Pi(\boldsymbol{\mathcal{G}_{m}},\boldsymbol{\mathcal{G}_{m}})} \sum_{u,v} T^{TT}_{uv}C^{TT}_{uv} + \lambda H(\hat{\mathbf{T}}).
\end{aligned}
\end{equation}
For $N$ image-caption pairs in a mini-batch, we compute the loss for each pair, resulting in $N$ losses $\{\mathcal{L}_i^s\}_{i=1}^{N}$. The total intra-modality fine-grained alignment loss is the sum of all individual losses as $\mathcal{L}_{SM} = \sum_{i=1}^N \mathcal{L}_i^s$. The alignment loss can separately enhance the intrinsic relationships of each modality, avoiding inter-modality fine-grained alignment compromises the intrinsic relationships of each modality.

Following RoCLIP~\cite{yang2023robust}, the caption pool is considered a first-in-first-out queue, which is initialized with random caption representations. After training on every mini-batch, we update this pool by taking the caption representations of the $N$ examples in the mini-batch and concatenating them at the end of the queue. We discard the oldest $N$ elements from the queue, which equals the training batch size.

\subsection{Training and Inference}
\noindent\textbf{Training.} 
To ensure the model performs well, we use a relatively large pool size for the image-caption pairs. This allows every clean image to find a caption that is similar to its original caption. To prevent the model from becoming overly focused on intra-modal features, we train using the intra-modal fine-grained alignment loss ($\mathcal{L}_{SM}$) every $K$ epochs. Follow RoCLIP~\cite{yang2023robust}, the $K$ is set to 2 during pre-training. The overall loss function can be formulated as follows:
\begin{equation}
\begin{aligned}
\small
\mathcal{L}_{\text{total}}
&= \lambda_c\mathcal{L}_c
+ \lambda_{IM}\mathcal{L}_{IM}\\
&+ \mathbf{1}\{\,\mathrm{epoch}\bmod K = 0\}\;\lambda_{SM}\mathcal{L}_{SM}.
\label{eq:loss}
\end{aligned}
\end{equation}
% \begin{equation}
% % \vspace{-2mm}
% \small
%    \mathcal{L}_{\text{total}} = 
%     \begin{cases} 
%         \lambda_c\mathcal{L}_c + \lambda_{SM}\mathcal{L}_{\text{SM}} + \lambda_{IM}\mathcal{L}_{\text{IM}}  & \text{ if } epoch \bmod K = 0, \\
%         \lambda_c\mathcal{L}_c + \lambda_{IM}\mathcal{L}_{\text{IM}} & \text{otherwise.} \\ 
%     \end{cases}
% \label{eq:loss}
% \end{equation}

\noindent\textbf{Inference.}  The global features are obtained by averaging the aligned fine-grained features. During inference, we follow previous methods to use the global features.

\begin{table*}[ht!]
\caption{Downstream linear probe and zero-shot (top-1) accuracy of pre-training on CC1M. Highest performance is bold, and the lowest is underscored. The last column highlights the average improvement over CLIP.}
\label{tab: Classification Utility}
\begin{center}
\begin{small}
\renewcommand{\arraystretch}{1.0} % Adjusts row spacing
\begin{tabular}{p{1.1cm}p{0.9cm}p{0.5cm}cccccccccc}
\toprule
Method & Task & F102 & Fd101 & I1K & Pet & Cars & Cal101  & C10 & C100 & DTD & Air. &Average\\
% Method & Task & \rotatebox[origin=c]{90}{F102} & \rotatebox[origin=c]{90}{Fd101} & \rotatebox[origin=c]{90}{I1K} & \rotatebox[origin=c]{90}{Pet} & \rotatebox[origin=c]{90}{Cars} & \rotatebox[origin=c]{90}{Cal101}  & \rotatebox[origin=c]{90}{C10} & \rotatebox[origin=c]{90}{C100} & \rotatebox[origin=c]{90}{DTD} & \rotatebox[origin=c]{90}{Air.} &\rotatebox[origin=c]{45}{\textbf{Alg-CLIP}}\\
\midrule
 & 0-shot & \underline{1.0}  & \textbf{7.1} & 9.6 & 3.4  & \textbf{0.8}  & 34.90   & 34.90  & 7.3 &  \underline{3.7} & 0.8 & 10.35 \\
CLIP & lin-prb & \underline{99.50}  & 44.90 & 22.20 & 48.20  & 12.90  & 70.40   & 70.50  & 45.80 & 48.20 & 24.90 & 48.75  \\
\midrule
 & 0-shot & 0.83  & 6.34 & 6.63 & 3.68  &0.72  & 30.38   & 30.14  & 9.52 & 3.56 & \textbf{1.11}& 9.291 \\ 
RoCLIP & lin-prb & 99.22  &\underline{54.05} & 24.09  & \underline{52.36} & \underline{20.35} & 72.15  & \underline{78.99}  & \underline{57.82}  & 55.21  & \underline{32.55} & \underline{54.679}\\
\midrule
 & 0-shot & 0.62  & 6.29 & \underline{9.87} & \textbf{5.51} & \underline{0.75} &  \underline{40.69}  & 39.7 & \underline{10.41} & 3.14 & 0.48 & \underline{11.746}\\
\alg & lin-prb &99.38  &  45.58 & \underline{24.53} & 51.02 & 15.35  & \underline{74.4} & 71.90  & 47.32  & \underline{56.01}  & 27.63 & 51.324   \\
\midrule
 & 0-shot & \textbf{1.19}  & \underline{6.57} & \textbf{10.50} & \underline{4.17} & 0.46 &  \textbf{45.38}  & \textbf{41.90} & \textbf{15.44} & \textbf{4.52} & \underline{0.99} & \textbf{13.112}\\
\textbf{\algo} & lin-prb &\textbf{99.81} &\textbf{56.26}  & \textbf{25.40} & \textbf{52.79} & \textbf{20.63} & \textbf{84.95}  & \textbf{79.17} & \textbf{58.46}  & \textbf{56.97}  &  \textbf{32.85} & \textbf{56.731}\\
\midrule
\bottomrule
\end{tabular}
\end{small}
\end{center}
\end{table*}
\setlength{\belowcaptionskip}{-3pt}

\section{Experimental Analyses}
\label{experience}

In this section, we evaluate the effectiveness of \algo against strong targeted data poisoning and backdoor attacks. We begin by outlining the experimental setup, followed by our main results, and conclude with an ablation study on various components of \algo.

\noindent\textbf{Pre-training Data.}  
To ensure broad dataset coverage, we utilize three diverse datasets: Conceptual Captions 3M (CC3M) \citep{sharma2018conceptual}, Visual Genome (VG) \citep{krishna2017visual}, and MSCOCO \citep{lin2014microsoft}. Following \citep{yang2023robust}, we randomly sample 1M image-caption pairs from CC3M (denoted as CC1M) to further evaluate \algo's defense capabilities. Throughout all experiments, we maintain a consistent set of hyperparameters: a learning rate of \(5 \times 10^{-5}\), \(\lambda_c = 1\), \(\lambda_{SM} = 0.4\), \(\lambda_{IM} = 2\), and $P$ = 10000. These settings demonstrate \algo's robustness against various types of attacks, independent of dataset distribution. Consistent with the setup in \citep{radford2021learning}, we employ a ResNet-50 as the image encoder and a transformer as the text encoder, training \algo from scratch over 32 epochs and the matching frequency is set to 2 to effectively counter the poison.

\noindent\textbf{Attack Baselines.} We follow the methodologies of previous work \citep{yang2023robust, yang2024better} to evaluate our defense strategy. For TDPAs, we randomly select images from the CC3M validation set as target images. Each target is assigned a random class from the ImageNet1K dataset \citep{deng2009imagenet}, and an adversarial caption set is constructed related to the adversarial label, as detailed in Sec. \ref{sec:pre_attack}. The poison rate is set at 0.05\% across all datasets. For BAs, we randomly select images from the CC3M validation set and apply the respective backdoor triggers. Each attack starts with a random class from the ImageNet1K dataset, creating adversarial caption sets. Each backdoor image pairs with a randomly chosen poisoned caption from this set. We evaluate with a poisoning ratio of 0.5\% for TPDA and 5\% for backdoor attacks on MSCOCO and Visual Genome. For CC1M, we use a 0.5\% poisoning ratio for both TPDA and the four additional backdoor attacks.

\subsection{Downstream Performance of \textbf{\algo}} \label{sec:res}\vspace{-1mm}
We evaluate the performance of \algo on several datasets from \cite{kornblith2019better}, with details provided in the Appendix.  It can be seen that effectively improves the zero-shot and linear-probe classification performance across all ten datasets in Table~\ref{tab: Classification Utility}. RoCLIP may introduce mismatching data by using CLIP’s global semantic matching, leading to a noticeable drop in zero-shot classification performance. To ensure the effectiveness of defense, \alg discards a large amount of clean data along with the poisoned samples, which reduces the model's linear probe performance. In contrast, \algo adopts a matching approach based on optimal transport to reconstruct image-caption pairs during pre-training. As a result, it avoids any decline in both zero-shot and linear probe classification performance, making our method more practical and effective.

\subsection{Defense Performance of \textbf{\algo}} \label{sec:res}\vspace{-1mm}
Here, we evaluate the performance of \algo against TDPA and BAs, comparing it with CLIP, RoCLIP, and \alg in terms of both ASR (Attack Success Rate) and downstream performance. Table \ref{tab:defense} demonstrates the high effectiveness of our \algo against CLIP, with ASRs exceeding 60\% for TDPA across all datasets and even surpassing 90\% for some BAs. This underscores the significant challenge in ensuring CLIP's robustness. In contrast, \algo significantly reduces the ASR to 0\% across all datasets for both TDPA and BAs. Although both RoCLIP and \alg provide decent defense, their performance is less consistent compared to \algo. For instance, \alg's ASR on some datasets is higher than that of \algo. \looseness=-1

\begin{table}[h]
\caption{Effectiveness of \algo in defending against various data poisoning attacks, measured by Attack Success Rate (ASR). \algo achieves a strong defense across datasets and attacks.}
\begin{small}
\setlength{\tabcolsep}{2pt} % Adjusts column spacing
\renewcommand{\arraystretch}{0.95} % Adjusts row spacing
\begin{tabular}{@{}lccccc@{}}
\toprule
\textbf{Dataset} & \multicolumn{5}{c}{\textbf{MSCOCO}} \\
\cmidrule(lr){2-6} 
Attacks & TDPA & BadNet & Label Consis & Blended & WaNet  \\
\midrule
CLIP            & 68.75\% & 31.0\% & 67.96\% & 92.50\% & 11.72\% \\
RoCLIP          & 43.75\%  & 5.63\% & 11.50\% & 43.60\% & 7.20\% \\
\alg       & 25\% & 0.33\% &  \textbf{0\%} & 36.67\% & 2.67\% \\
\textbf{\algo}      & 6.25\% & \textbf{0\%} &  \textbf{0\%} & \textbf{0\%} & \textbf{0\%}  \\
\midrule
\textbf{Dataset} & \multicolumn{5}{c}{\textbf{Visual Genome}}  \\
\cmidrule(lr){2-6} 
Attacks & TDPA & BadNet & Label Consis & Blended & WaNet  \\
\midrule
CLIP            & 75.00\% & 6.90\% & 32.84\% & 86.97\% & 19.96\% \\
RoCLIP          & 37.5\% & 4.33 & 7.31\% & 19.60\% &9.71\% \\
\alg       & 6.25\% & \textbf{0\%} & \textbf{0\%} & 6.33\% & \textbf{0\%}  \\
\textbf{\algo}      & \textbf{0\%} & \textbf{0\%} & \textbf{0\%} & \textbf{0\%} & \textbf{0\%} \\
\midrule
\textbf{Dataset} & \multicolumn{5}{c}{\textbf{CC1M}}  \\
\cmidrule(lr){2-6} 
Attacks & TDPA & BadNet & Label Consis & Blended & WaNet  \\
\midrule
CLIP            & 93.75\% & 93.25\% & 71.0\% & 99.30\% & 97.42\%  \\
RoCLIP          & 56.25\% & 11.72\% & 5.31\% & 23.71\%& 26.27\%  \\
\alg       & 6.25\% & \textbf{0\%} & \textbf{0\%} & 5.47\% & 3.43\%  \\
\textbf{\algo}      & \textbf{0\%} & \textbf{0\%} & \textbf{0\%} & 0.3\% & \textbf{0\%} \\
\midrule
\bottomrule
\end{tabular}
\label{tab:defense}
\end{small}
\end{table}
\setlength{\belowcaptionskip}{-2pt}

% \begin{table}[h]
% \caption{\algo is more effective in defending against poisoning attacks compared to the previous state-of-the-art method (\alg) under higher attack rates.}
% \vspace{-2mm}
% \begin{small}
% \setlength{\tabcolsep}{4pt} % Adjusts column spacing
% \renewcommand{\arraystretch}{0.8} % Adjusts row spacing
% \begin{tabular}{@{}lccccc@{}}
% \toprule
% \textbf{Dataset} & \multicolumn{5}{c}{\textbf{MSCOCO}} \\
% \cmidrule(lr){2-6} 
% Attacks & TDPA & BadNet & Label Consis & Blended & WaNet  \\
% \midrule
% \alg       & 25\% & 0.33\% & \textbf{0\%} & 36.67\% & 2.67\% \\
% \textbf{\algo}        & \textbf{0\%} & \textbf{0\%} & \textbf{0\%} & \textbf{0\%} & \textbf{0\%} \\
% \midrule
% \textbf{Dataset} & \multicolumn{5}{c}{\textbf{Visual Genome}}  \\
% \cmidrule(lr){2-6} 
% Attacks & TDPA & BadNet & Label Consis & Blended & WaNet  \\
% \midrule
% \alg       & 6.25\% & \textbf{0\%} & \textbf{0\%} & 6.33\% & \textbf{0\%}  \\
% \textbf{\algo}      & \textbf{0\%} & \textbf{0\%} & \textbf{0\%} & \textbf{0\%} & \textbf{0\%} \\
% \midrule
% \textbf{Dataset} & \multicolumn{5}{c}{\textbf{CC1M}}  \\
% \cmidrule(lr){2-6} 
% Attacks & TDPA & BadNet & Label Consis & Blended & WaNet  \\
% \midrule
% \alg       & 6.25\% & \textbf{0\%} & \textbf{0\%} & 5.47\% & 3.43\%  \\
% \textbf{\algo}      & \textbf{0\%} & \textbf{0\%} & \textbf{0\%} & 0.3\% & \textbf{0\%} \\
% \bottomrule
% \end{tabular}
% \label{tab:attack_higher}
% \end{small}
% \vspace{-6mm}
% \end{table}
% \setlength{\belowcaptionskip}{-2pt}

\subsection{Ablation Study}
% \noindent\textbf{Attacks With Higher Poisoning Rate.}\label{sec:higer_attack} 
% Our method has demonstrated effectiveness with a TPDA poisoning rate of 0.05\%, BadNet at 0.05\%, and other backdoor methods at 0.15\%. However, compared to the previous state-of-the-art method, SafeCLIP, our approach does not exhibit a significant advantage at lower poisoning ratios. To address this, we explore higher poisoning ratios: 0.5\% for TPDA and 5\% for backdoor attacks on MSCOCO and Visual Genome. We apply a poisoning ratio of 0.5\% for TPDA and the four other backdoor attacks on CC1M. We maintain the same hyperparameter settings as in Section~\ref{experience}. As shown in Table \ref{tab:attack_higher}, \algo demonstrates greater effectiveness against poisoning data attacks at these higher poisoning ratios.\looseness=-1

\noindent\textbf{Impact of Optimal Transport-based Matching.} We conducted ablation experiments to evaluate the impact of Optimal Transport Matching. As shown in Table \ref{ablation}, replacing  optimal transport-based matching with CLIP's semantic matching significantly improves ASR across all datasets and decreases CLIP's zero-shot and linear probing performance. This highlights the importance of Optimal Transport Matching in constructing clean samples. \looseness=-1

\noindent\textbf{Impact of Inter-modality Fine-grained Alignment.} The third row of Table \ref{ablation} shows that removing the inter-modality fine-grained alignment leads to a decrease in CLIP's zero-shot and linear probing performance. This demonstrates that inter-modality fine-grained alignment is essential for better aligning the fine-grained features of image-caption pairs and improving generalization performance. \looseness=-1

\noindent\textbf{Impact of Intra-modality Fine-grained Alignment.} We also evaluate the impact of intra-modality fine-grained alignment. From Table~\ref{ablation}, we can observe that removing the content relationship within each modality leads to a decrease in CLIP's zero-shot and linear probing performance. This proves that intra-modality fine-grained alignment is helpful in improving model performance. \looseness=-1

\begin{table}[h]
    \centering
    \renewcommand{\arraystretch}{1.0} % Adjusts row spacing
    \caption{Ablation study for different module. Linear probe and zero-shot performance is reported on CIFAR-10 (C10), CIFAR-100 (C100), ImageNet-1K (I1K).}
    \resizebox{\linewidth}{!}{
        \begin{threeparttable}
            \begin{tabular}{ccc|c|c|c|c|c}
                \toprule
                \textcircled{1} &  \textcircled{2} & \textcircled{3} & Task & C10 & C100 & I1K & TDPA  \\ 
                \midrule
                \multirow{2}{*}{\ding{51}} & \multirow{2}{*}{\ding{51}} & \multirow{2}{*}{\ding{51}} & 0-shot & 41.90 & 15.44 & 10.50 & \multirow{2}{*}{ 0\%}  \\
                & & & lin-prb & 79.17 & 58.46 & 25.40 \\
                \midrule
                \multirow{2}{*}{\ding{56}} & \multirow{2}{*}{\ding{51}} & \multirow{2}{*}{\ding{51}} & 0-shot & 36.47 & 11.30 & 8.60 & \multirow{2}{*}{12.5\%}  \\
                & & & lin-prb & 75.03 & 55.90 & 23.19  \\
                \midrule
                \multirow{2}{*}{\ding{51}} & \multirow{2}{*}{\ding{56}} & \multirow{2}{*}{\ding{51}} & 0-shot  & 39.62 & 13.60 & 9.70  & \multirow{2}{*}{0\%}   \\
                & & & lin-prb & 76.50 & 56.80 & 24.10   \\
                \midrule
                \multirow{2}{*}{\ding{51}} & \multirow{2}{*}{\ding{51}} & \multirow{2}{*}{\ding{56}} & 0-shot & 40.15 & 14.17 & 10.63 & \multirow{2}{*}{0\%}  \\
                & & & lin-prb   & 78.10 & 57.23 & 23.37 \\
                \bottomrule
            \end{tabular}
            \begin{tablenotes}[flushleft] % Set left alignment for tablenotes
                \small
                \item \textcircled{1} Optimal Transport-based Matching (section~\ref{sec:inter_match})
                \item \textcircled{2} Inter-modality Fine-grained Alignment (section~\ref{sec:inter})
                \item \textcircled{3} Intra-modality Alignment (section~\ref{sec:intra})
            \end{tablenotes}         
        \end{threeparttable}
    }
    \label{ablation}
    \vspace{-2mm}
\end{table}

\subsection{Visualization of \textbf{\algo} Matching}
\begin{figure}[t!]
    \begin{center}
        \includegraphics[width=1\linewidth]{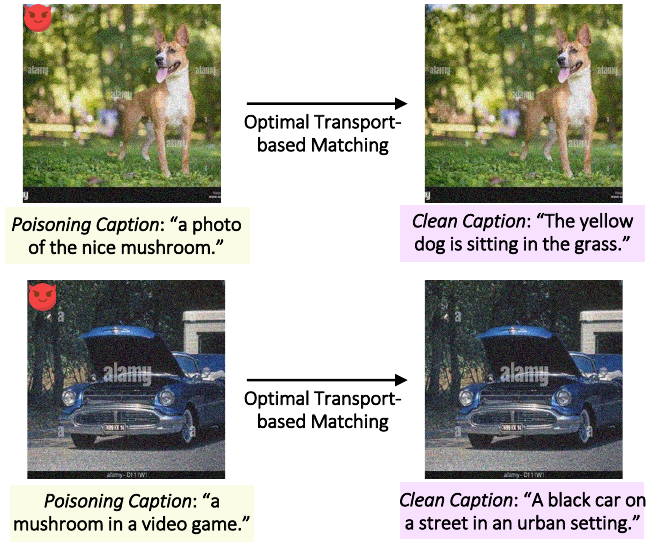}
    \end{center}
    \caption{Visualization results of \algo re-matching to the most similar caption in caption pool based on optimal transport.}
    \label{fig:fig_vis}
    \vspace{-2mm}
\end{figure}

As shown in Figure~\ref{fig:fig_vis}, when poisoned data is fed into \algo, \algo first breaks the association between poisoned image-caption pairs and then re-matches each image to a caption that is most similar. In Figure~\ref{fig:fig_vis}, we can see that optimal transport-based matching effectively matching captions with semantics similar to the images.

\vspace{-1mm}

\section{Conclusion}
Recent studies have shown that  CLIP is extremely vulnerable to targeted data poisoning and backdoor attacks. Previous methods solely rely on the global representations of images and captions, overlooking fine-grained features. To address their limitations, we propose an Optimal Transport-based framework to reconstruct the image-caption pairs, named \algo. It models images and captions with fine-grained visual and textual feature sets and re-assigns new captions based on optimal transport distance. Additionally, we encourage the inter- and intra-modality fine-grained alignment by employing optimal transport-based objective functions. Our experiments demonstrate that \algo can successfully decrease the attack success rates. Compared to previous methods, \algo significantly improves CLIP's zero-shot and linear probing performance trained on poisoned datasets. 

\section*{Limitations}
We employ Optimal Transport-based matching to defend against data poisoning and backdoor attacks. However, we note that although the model's defense performance has improved, the need for Sinkhorn iterations to compute the optimal transport matrix introduces additional computational overhead. These iterations require more time and computational resources compared to directly utilizing CLIP’s similarity-based computations. While this trade-off enhances defense effectiveness, the increased resource consumption may become a limiting factor, particularly in large-scale defense scenarios with extensive datasets. We acknowledge this limitation and plan to optimize the OT-based process in future work to reduce computational cost and improve overall efficiency without compromising defense performance.

\section*{Ethics Statement}
While the malicious application of data poisoning and backdoor attacks may raise ethical concerns, we propose a more effective defense method using Optimal Transport to mitigate these threats. This approach can help minimize potential harm from such vulnerabilities. The primary goal of this work is to encourage the development of appropriate defense mechanisms rather than to promote malicious use. We believe that by addressing these challenges, our efforts will inspire the research community to create more responsible and secure AI systems, fostering the development of trustworthy models that can better withstand adversarial attacks.

\section*{Acknowledgement}
This work is supported in part by the Key Research and Development Program of Zhejiang Province under Grant No. 2025C02103.

\bibliography{acl_latex}
\appendix
\section{Appendix}
\subsection{Experimental Setup}
\label{sec:add_expr}
\subsubsection{Training Dataset}
\noindent\textbf{MSCOCO.} MSCOCO \citep{lin2014microsoft} is a large-scale dataset designed for object detection, segmentation, and captioning. It includes 80 object categories, with each image paired with 5 captions. For our analysis, we randomly select one caption per image, resulting in a dataset size of 80K images.

\noindent\textbf{Visual Genome.} Visual Genome \citep{krishna2017visual} is an extensive dataset focused on region captions. It contains 10,877 images and 5.4 million region descriptions. For each image, we randomly select 5 region descriptions and combine them into a single caption.

\noindent\textbf{Conceptual Captions.} Conceptual Captions \citep{sharma2018conceptual} is a large-scale, web-based image captioning dataset that covers a diverse range of image styles and caption formats.

\begin{algorithm}[ht]
\small
  \caption{\algo for Defense Against Data Poisoning}\label{alg:enhanced_alg}
  \begin{algorithmic}[1]
    \State \textbf{Input}: 
    \begin{itemize}
      \item Image encoder $E_I$, text encoder $E_T$
      \item \algo frequency $K$
      \item Fine-grained caption pool $\mathcal{P}^s = \{y^s_{p(i)}\}_{i=1}^P$ and global caption pool $\mathcal{P}^g = \{y^g_{p(i)}\}_{i=1}^P$, initialized with random captions
    \end{itemize}
    \For{epoch = $1, \ldots, T$}
      % \For{each mini-batch of image-caption pairs $\{(X_i, Y_i)\}_{i=1}^N \in D$ with corresponding fine-grained features $(f^s_i, y^s_i)_{i=1}^N$ and global features $(f^g_i, y^g_i)_{i=1}^N$}
      \For{each mini-batch of image-caption pairs 
    \hspace*{3em} $\{(X_i, Y_i)\}_{i=1}^N \in D$ with corresponding 
    \hspace*{4em} fine-grained features $(f^s_i, y^s_i)_{i=1}^N$ and global 
    \hspace*{4em} features $(f^g_i, y^g_i)_{i=1}^N$}

        \If{$\text{epoch} \bmod K == 0$}
        \State \textcolor[rgb]{0.165, 0.49, 0.47}{//\textbf{optimal transport-based matching score}}
          \For{$i = 1, \ldots, N$}
            \State $M = \min_{\mathbf{T} \in \Pi(\mathcal{G}_f, \mathcal{G}_p)} \sum_{i,j} T_{ij} C_{ij}^P + $
            \State $\lambda H(\mathbf{T})$
            \State $\hat{M} = 1 - M$
          \EndFor
            \State $\mathcal{M} = \{\hat{M}_i\}_{i=1}^N$
            \State \textcolor[rgb]{0.165, 0.49, 0.47}{//\textbf{extract the indices of the best matches}}
            \State  $\text{Index} = \arg\max_{i} \mathcal{M}[:,i]$
            \State \textcolor[rgb]{0.165, 0.49, 0.47}{//\textbf{Retrieve updated captions:}}
            \State  $y^{s}_{m} = y^{s}[:, \text{Index}]$
            \State  $y^{g}_{m} = y^{g}[:, \text{Index}]$
            \State \textcolor[rgb]{0.165, 0.49, 0.47}{//\textbf{train encoders with loss:}}
            \State  $\mathcal{L} = \lambda_c\mathcal{L}_c + \lambda_{\text{SM}}\mathcal{L}_{\text{SM}} + \lambda_{\text{IM}}\mathcal{L}_{\text{IM}}$
        \Else
          \State \textcolor[rgb]{0.165, 0.49, 0.47}{//\textbf{train encoders with simplified loss}}
          \State  $\mathcal{L} = \lambda_c\mathcal{L}_c + \lambda_{\text{IM}}\mathcal{L}_{\text{IM}}$
        \EndIf
      \EndFor
    \EndFor
  \end{algorithmic}
\end{algorithm}

\begin{algorithm}[ht]
\small
\caption{Sinkhorn Iteration for Optimal Transport}
\label{alg:sinkhorn}
\begin{algorithmic}[1]
\Require $C$: cost matrix, $P$: number of caption pool, $h \times w$: number of spatial image features, $l$: length of caption, $\beta$: scaling parameter
\Ensure $T$: transport matrix
\State $\sigma \gets \text{ones\_like}(P, h \times w, 1) / m$
\State $T \gets \text{ones\_like}(P, l, h \times w)$
\State $A \gets \exp(-(\text{clamp}(C, \max(10 \cdot \beta))) / {\beta})$
\For{$i = 1$ \textbf{to} $100$}
    % \State $r_0 \gets r$
    \State $\delta \gets 1 / 1 / {n \cdot \sum(Q \cdot \sigma, \text{ axis}=2)}$
    \State $a = \sum(Q \cdot \delta, \text{ axis}=2)$
    \State $\sigma = 1 / {m \times a}$
    \State $T \gets \delta \times Q \times K$
\EndFor
\State \Return $T$
\end{algorithmic}
\end{algorithm}
\subsubsection{Evaluation Setup for Targeted Data Poisoning}
\noindent\textbf{Downstream Dataset.}
To assess the downstream performance of our model, we perform linear probing and zero-shot classification, as detailed in Sec.~\ref{sec:pre_clip}, on 10 widely adopted datasets \citep{radford2021learning,li2021supervision,yang2023robust} listed in Table~\ref{tab:downstream_datasets}.

\begin{table}[h!]
\caption{Details of downstream datasets.}
\label{tab:downstream_datasets}
% \vskip 0.15in
\vspace{-2mm}
\begin{center}
\begin{small}
% \begin{sc}
% \resizebox{!}{0.3\linewidth}{
\begin{tabular}{llrr}
\toprule
\textbf{Dataset} & \textbf{Classes} & \textbf{Train Size} & \textbf{Test Size} \\
\midrule
CIFAR10 & 10 & 50,000 & 10,000\\
CIFAR100 & 100 & 50,000 & 10,000\\
Food-101 & 101  & 75,750 & 25,250\\
DTD & 47 & 3,760 & 1,880 \\ 
FGVC Aircraft & 100 & 6,667 & 3,333 \\
Flowers-102 & 102 &  2,040 & 6,149\\
Caltech-101 & 102 & 3,060 & 6,085\\
OxfordIIITPet & 37  & 3,680 & 3,669\\
Stanford Cars & 196 & 8,144 & 8,041\\
ImageNet1K & 1000  & 50,000 & 50,000\\
% Flickr30K & - & 31,783 & - \\
\bottomrule
\end{tabular}
% }
% \end{sc}
\end{small}
\end{center}
\vspace{-2mm}
\end{table}
\noindent\textbf{Zero-shot Classification.} 
Zero-shot classification assesses the generalization and transferability of the model to unseen tasks. It transforms the downstream labels into natural language captions using the provided engineered prompt templates, such as "\texttt{A photo of a \{label\}}" \citep{radford2021learning}. Then, it calculates the cosine similarity between the representations of a given image and each prompt and predicts the label with the highest image-prompt similarity.\looseness=-1

\noindent\textbf{Linear Probe Classification.} 
Linear probe classification refers to evaluating the extracted representations from the pre-trained image encoder to train a linear classifier on the downstream labeled data. 

\subsubsection{Defense Baselines for Backdoor} We consider RoCLIP~\citep{yang2023robust} and \alg~\cite{yang2024better} as our baseline. We measure the effectiveness of attacks using attack success rate (ASR). For TDPA, ASR is measured as the fraction of target images that are classified as the adversarial label. For BA, ASR is measured as the fraction of test images containing the backdoor triggers that are classified as the adversarial label.

\subsubsection{Backdoor Attacks Used in Our Evaluations}
\begin{figure}[ht]
    \centering
    \begin{subfigure}{.32\columnwidth}
        \centering
        \includegraphics[width=\linewidth]{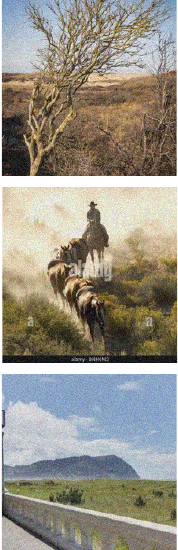}
        \caption{Blended}
    \end{subfigure}
    \hfill
    \begin{subfigure}{.32\columnwidth}
        \centering
        \includegraphics[width=\linewidth]{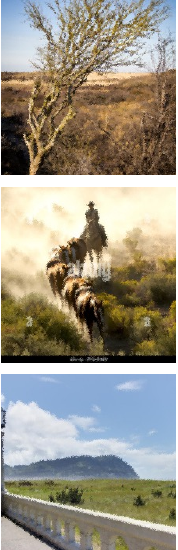}
        \caption{WaNet}
    \end{subfigure}
    \hfill
    \begin{subfigure}{.32\columnwidth}
        \centering
        \includegraphics[width=\linewidth]{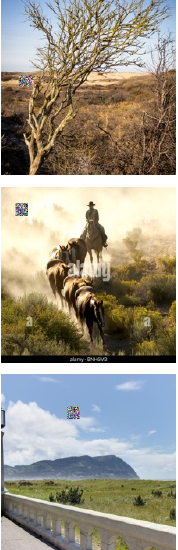}
        \caption{BadNets}
    \end{subfigure}
    \caption{Backdoor attacks used in our evaluations.}
    \label{fig:attack}   
\end{figure}

We follow the methodologies of previous work \citep{yang2023robust, yang2024better} to evaluate our defense strategy against backdoor attacks (BA) with visible triggers (e.g., BadNet) and invisible triggers (e.g., Blended and WaNet). Figure~\ref{fig:attack} illustrates various examples of backdoor attacks for visualization.

\subsubsection{Hyperparameters Setting}

The hyperparameter settings used in our experiments are provided in Table \ref{tab:tuning}. As shown in Table \ref{tab:tuning}, our model employs consistent hyperparameters across all datasets, highlighting the robustness of \algo against a variety of attacks.

\begin{table}[h]
\centering
\caption{{Hyperparameters of our experiments}.}\label{tab:tuning}
\begin{tabular}{l|l|l|l}
\toprule
\textbf{Dataset}  & \textbf{lr} & \textbf{Batch Size} &\textbf{K} \\ 
\midrule
CC3M & 5e-5  & 256 & 2 \\
CC1M & 5e-5  & 256 & 2\\
COCO & 5e-5  & 256 & 2\\
VG   & 5e-5  & 256 & 2\\
\bottomrule
\end{tabular}
\end{table}

\subsection{\textbf{\algo} Against Adaptive Attacks}
In the above experiments, we assume that attackers have no information about our backdoor defense. In this section, we consider a more challenging setting, where the attackers know the existence of our defense and can construct the
poisoned dataset with an adaptive attack.

\noindent\textbf{Threat Model For The Attackers.} Following existing work~\cite{gao2023backdoor}, we assume that the attackers can access all dataset and know the architecture of the victim model. However, the attackers can not control the training process
after poisoned samples are injected into the training dataset.

\noindent\textbf{Methods.} Our defense method uses optimal transport-based matching to separate samples and reconstruct image-caption pairs. For adaptive attacks, the goal is to minimize the difference in optimal transport-based matching between the image and its poisoned caption. Attackers first use the image encoder and text encoder to extract spatial and token sequence features from the poisoned pairs. Then, the loss function defined in Eq.~\ref{eq:matching} is applied to update the trigger patch. This pattern is optimized by minimizing the gradient of the poisoned image-caption pair, reducing the distance between them in the fine-grained feature space.

\noindent\textbf{Settings.} We conduct experiments on the poisoning image-caption pairs. We adopt projected gradient descent (PGD)~\cite{wang2018analyzing} to optimize the trigger pattern for 100 iterations.

\noindent\textbf{Results.} This adaptive attack achieves a 0\% attack success rate on both MSCOCO and Visual Genome, demonstrating that our defense effectively resists adaptive attacks. \looseness=-1

\subsection{Additional Experiments}
\subsubsection{\algo's Complexity Compared to Existing Defense Methods}\label{sec:overhead} 
RoCLIP leverages CLIP's global employ global feature vectors to match the most similar for every image, aiming to break the association between poisoned pairs. \alg identifies the clean and risky set using global features. \alg apply the CLIP loss
to the safe set and \alg apply unimodal CL to image and text modalities of the risky set separately. \alg performs data augmentation on the risky data and applies unimodal contrastive learning (CL) in the risky and augmented data. We propose the optimal transport-based fine-grained matching and alignment against data poisoning.

As shown in the Table~\ref{tab:training_time}, we calculated the computational cost of these three methods within a single epoch under the same settings. From the methodology section, we found that \algo requires using Sinkhorn iteration~\cite{distances2013lightspeed} to obtain the optimal transport matrix. As shown in Table~\ref{tab:training_time}, we calculated the computational cost of these three methods per epoch under the same settings. According to the methodology, \algo requires Sinkhorn iteration~\cite{distances2013lightspeed} to compute the optimal transport matrix, introducing slightly more computational time compared to RoCLIP. However, it is significantly faster than \alg.
As noted in \alg~\cite{yang2024better}, data augmentation is applied to risky data, generating augmented samples. Both the augmented and original risky data are used for training. Since approximately 75\% of the data are marked as risky data, the training data set almost doubles in size, significantly increasing the training time.

\begin{table}[ht]
\centering
\caption{Training time of \algo compared to existing defense methods.}
\label{tab:training_time}
\begin{small}
\setlength{\tabcolsep}{8pt}
\begin{tabular}{ll}
\hline
\textbf{Method} & \textbf{Training Time} \\
\hline
RoCLIP & 1 h 23 min \\
\alg & 4 h 11 min \\
\algo & 2 h 7 min \\
\hline
\end{tabular}
\end{small}
\end{table}

% \subsection{Hyperparameter Tuning}
% We include the hyperparameter settings of our experiments in Table \ref{tab:tuning}. There are few key hyperparameters for tuning:

% \textbf{\alg with different model architecture} CLIP has two variations in its vision model, ResNet, which we used in our original experiments, and ViT, which we report here. We attack the model with TDPA and BadNet backdoor with a poison rate of 0.05\%, consistent with the setting in the paper. As shown in Table \ref{tab:model_poison_rate}, in both architectures, \alg can defend the model with the same hyperparameter setting.

% \begin{algorithm}[t]
% \caption{Sinkhorn Iteration for Optimal Transport}
% \label{alg:sinkhorn}
% \begin{algorithmic}[1]
% \Require $C$: cost matrix, $P$: number of caption pool, $h \times w$: number of spatial image features, $l$: length of caption, $\beta$: scaling parameter
% \Ensure $T$: transport matrix

% \State $\sigma \gets \text{ones\_like}(P, h \times w, 1) / m$
% \State $T \gets \text{ones\_like}(P, l, h \times w)$
% \State $A \gets \exp(-(\text{clamp}(C, \max(10 \cdot \beta))) / {\beta})$
% \For{$i = 1$ \textbf{to} $100$}
%     % \State $r_0 \gets r$
%     \State $\delta \gets 1 / 1 / {n \cdot \sum(Q \cdot \sigma, \text{ axis}=2)}$
%     \State $a = \sum(Q \cdot \delta, \text{ axis}=2)$
%     \State $\sigma = 1 / {m \times a}$
%     \State $T \gets \delta \times Q \times K$
% \EndFor
% \State \Return $T$
% \end{algorithmic}
% \end{algorithm}
\subsubsection{More Ablation Studies}
\noindent\textbf{The Sensitivity of The Caption Pool.} 
Next, we analyze the effect of pool size on our method. We apply \algo with pool sizes of 1\%, 2\%, and 10\% of the pre-training dataset. As shown in Table~\ref{ablation_size}, the pool size does not significantly impact the effectiveness of the defense. Across different pool sizes, our method consistently defends against data poisoning attacks. However, a larger pool size improves the downstream performance of the model, as it increases the likelihood of images finding more suitable captions.

\begin{table}[h]
    \centering
    \renewcommand{\arraystretch}{0.7} % Adjusts row spacing
    \caption{The impact of caption pool size. Linear probe and zero-shot performance is reported on CIFAR-10 (C10), CIFAR-100 (C100), ImageNet-1K (I1K).}
    \small % Change the font size for the table
    \resizebox{\linewidth}{!}{
        \begin{threeparttable}
            \begin{tabular}{c|c|c|c|c|c}
                \toprule
                Pool Size & Task & C10 & C100 & I1K & TDPA  \\ 
                \midrule
                 \multirow{2}{*}{1024} & 0-shot & 41.50 & 15.20 & 9.6 & \multirow{2}{*}{0\%}  \\
                 & lin-prb & 79.03 & 55.7 & 24.3  \\
                \midrule
                \multirow{2}{*}{2048} & 0-shot & 40.13 & 15.07 & 10.76 & \multirow{2}{*}{0\%}   \\
                 & lin-prb & 79.40 & 56.64 & 24.9  \\
                \midrule
                  \multirow{2}{*}{10000} & 0-shot & 41.90 & 15.44 & 10.50 & \multirow{2}{*}{0\%}  \\
                 & lin-prb & 79.17 & 58.46 & 25.40  \\
                \bottomrule
            \end{tabular}
  
        \end{threeparttable}
    }
    \label{ablation_size}
\end{table}

\subsubsection{Impact of different learning rates and loss function.}
We conducted ablation experiments for loss function. From the table below, we can see that CLIP Loss $\lambda_c$ is crucial to maintaining CLIP's performance. Inter-modal fine-grained alignment Loss $\lambda_{IM}$ is essential to better align the fine-grained features of image-caption pairs and improve generalization performance. Intra-modal fine-grained alignment loss $\lambda_{SM}$ is also helpful in improving the performance of the model.
\begin{table}[h]
    \centering
    \renewcommand{\arraystretch}{0.9} % Adjusts row spacing
    \caption{Ablation study for different loss functions. TPDA, linear probe, and zero-shot performance are reported on CIFAR-10 (C10), CIFAR-100 (C100), ImageNet-1K (I1K).}
    \resizebox{\linewidth}{!}{
        \begin{threeparttable}
            \begin{tabular}{c|c|c|c|c|c}
                \toprule
                Loss Function & Task & C10 & C100 & I1K & TDPA (\%) \\ 
                \midrule
                \multirow{2}{*}{ALL Loss} & 0-shot & 41.90 & 15.44 & 10.50 & \multirow{2}{*}{0} \\
                                         & lin-prb & 79.19 & 58.46 & 25.40 & \\
                \midrule
                \multirow{2}{*}{\textit{w/o.} $\lambda_c$} & 0-shot & 32.70 & 8.40 & 7.39 & \multirow{2}{*}{25.0} \\
                                                          & lin-prb & 72.12 & 47.35 & 21.51 & \\
                \midrule
                \multirow{2}{*}{\textit{w/o.} $\lambda_{IM}$} & 0-shot & 39.62 & 13.60 & 9.70 & \multirow{2}{*}{0} \\
                                                           & lin-prb & 76.50 & 56.80 & 24.10 & \\
                \midrule
                \multirow{2}{*}{\textit{w/o.} $\lambda_{SM}$} & 0-shot & 40.15 & 14.17 & 10.63 & \multirow{2}{*}{0} \\
                                                           & lin-prb & 78.10 & 57.23 & 23.37 & \\
                \bottomrule
            \end{tabular} 
        \end{threeparttable}
    }
    \label{ablation}
\end{table}

The table below presents an ablation study investigating the effect of different learning rates on model performance. The results indicate that a learning rate of $5e-5$ achieves the most favorable balance, yielding the highest zero-shot and linear probe accuracies across all datasets, while simultaneously reducing the attack success rate to 0\%. By contrast, lower or higher learning rates result in reduced predictive performance and increased vulnerability to poisoning, as reflected by increased TPDA values.

\subsection{Additional Visualization}
\begin{figure}[ht]
    \begin{center}
        \includegraphics[width=1.0\linewidth]{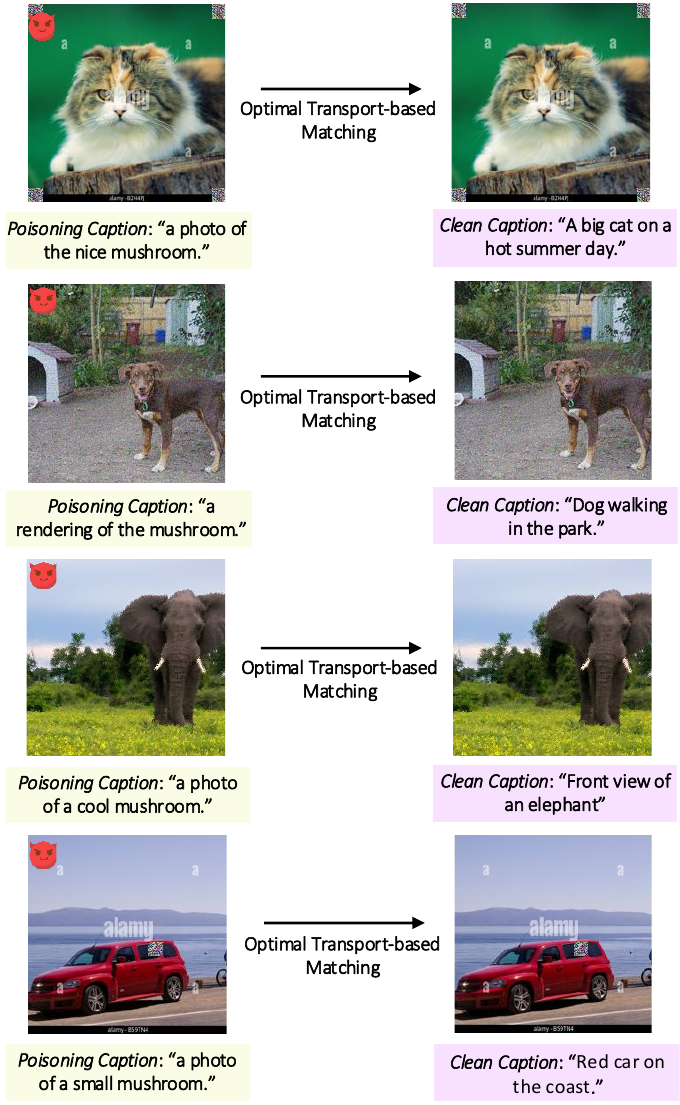}
    \end{center}
    \caption{Visualization results of \algo re-matching to the most similar caption in caption pool based on optimal transport.}
    \vspace{-2mm}
    \label{fig:fig_vis_appendix}
\end{figure}
As illustrated in Figure~\ref{fig:fig_vis_appendix}, we provide the matching results in various attack scenarios. The results indicate that \algo can disrupt the associations between the poisoned image-caption pairs and reassign each image to the most semantically compatible caption. Moreover, Figure~\ref{fig:fig_vis} further demonstrates that the optimal transport-based matching strategy effectively aligns images with captions that are semantically consistent, even in the presence of adversarial perturbations.

\end{document}